\documentclass[sensors,article,submit,pdftex,moreauthors]{Definitions/mdpi} 

\firstpage{1} 
\pubvolume{1}
\issuenum{1}
\articlenumber{0}
\pubyear{2026}
\copyrightyear{2026}
\datereceived{ } 
\daterevised{ } 
\dateaccepted{ } 
\datepublished{ } 

\usepackage{listings}
\usepackage{xcolor}

\definecolor{codeblue}{rgb}{0.25,0.5,0.5}

\lstdefinestyle{mdpipython}{
    language=Python,
    basicstyle=\small\ttfamily\selectfont,
    breaklines=true,
    commentstyle=\color{codeblue},
    keywordstyle=\color{black},
    backgroundcolor=\color{white},
    captionpos=b,
}
\Title{CoRe: Joint Optimization with Contrastive Learning for Medical Image Registration}

\Author{Eytan Kats, Christoph Grossbroehmer, Ziad Al-Haj Hemidi, Fenja Falta, Wiebke Heyer and Mattias P. Heinrich*}

\longauthorlist{yes}

\AuthorNames{Eytan Kats, Christoph Grossbroehmer, Ziad Al-Haj Hemidi, Fenja Falta, Wiebke Heyer and Mattias Heinrich}

\address{%
Insitute of Medical Informatics, University of Luebeck, Luebeck, Germany
}

\corres{Correspondence: mattias.heinrich@uni-luebeck.de}

\abstract{
Medical image registration is a fundamental task in medical image analysis, enabling the alignment of images from different modalities or time points. However, intensity inconsistencies and nonlinear tissue deformations pose significant challenges to the robustness of registration methods. Recent approaches leveraging self-supervised representation learning show promise by pre-training feature extractors to generate robust anatomical embeddings, that farther used for the registration. In this work, we propose a novel framework that integrates equivariant contrastive learning directly into the registration model. Our approach leverages the power of contrastive learning to learn robust feature representations that are invariant to tissue deformations. By jointly optimizing the contrastive and registration objectives, we ensure that the learned representations are not only informative but also suitable for the registration task. We evaluate our method on abdominal and thoracic image registration tasks, including both intra-patient and inter-patient scenarios. Experimental results demonstrate that the integration of contrastive learning directly into the registration framework significantly improves performance, surpassing strong baseline methods.
}


\begin{document}

\section{Introduction}
\label{sec:intro}

\begin{figure*}[t]
   \includegraphics[width=1.0\linewidth]{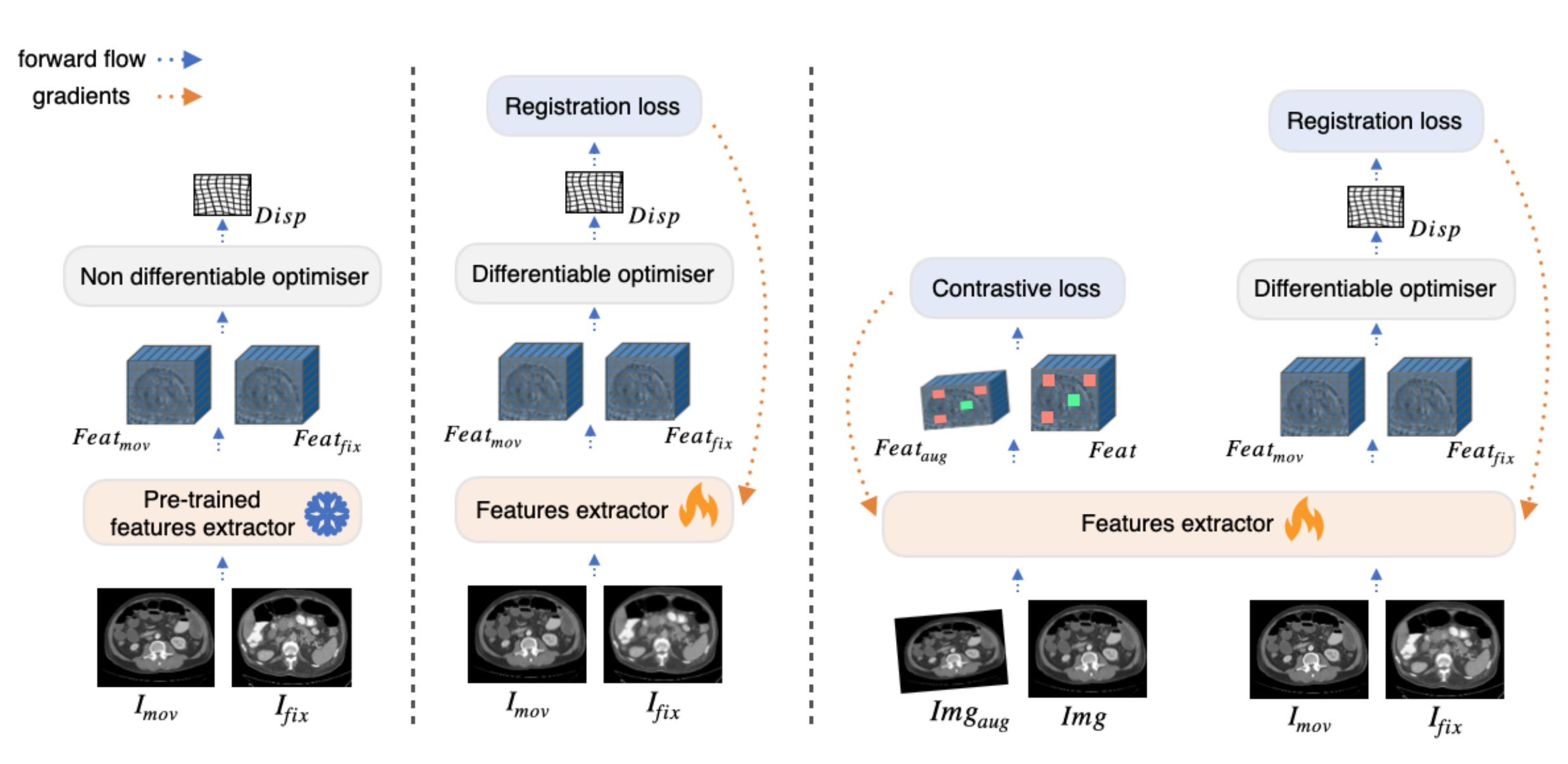}

   \caption{Comparison of hybrid registration methods. From left to right: (1) Feature extractor pretrained separately and used without further optimization during registration (SAMConvex \cite{li2023samconvex}); (2) Feature extractor optimized exclusively with a registration loss during training (Bigalke et~al. \cite{bigalke2023unsupervised}); (3) Proposed CoRe method, where the feature extractor is jointly optimized under both registration and contrastive loss objectives to enhance feature robustness and registration accuracy.}
   \label{fig:highlight}
\end{figure*}

Medical image registration is a fundamental problem in medical image analysis, aiming to establish dense anatomical and semantic correspondences between images. These images may be acquired at different time points, from different subjects, or using different imaging modalities. Accurate registration is a prerequisite for a wide range of downstream clinical and research tasks. It enables clinicians and researchers to track disease progression over time, evaluate the effectiveness of therapeutic interventions, quantify structural or functional changes, and analyze anatomical variability across patient populations.

The complexity of medical image registration arises primarily from two key factors: variability in image appearance and nonlinear anatomical deformations. Intensity variations commonly occur across scans due to differences in acquisition protocols, scanner hardware, or imaging modalities, making direct pixel-wise comparisons unreliable. Non-rigid deformations in anatomical structures are frequently observed in longitudinal scans of the same patient and in inter-subject comparisons within population studies. These deformations may also result from pathological changes or surgical interventions, adding further complexity to the registration process. To address these challenges, considerable research has focused on developing robust registration algorithms that can handle appearance variability while accurately modeling complex, spatially varying deformations.

\textbf{Mutual information}: The seminal work by Maes et al. \cite{maes1997multimodality} introduced mutual information (MI) as a similarity measure for the registration of multimodal medical images. Unlike traditional intensity-based metrics that rely on direct pixel value comparisons, MI quantifies the statistical dependence between images, enabling alignment based on their overall information content. Nevertheless, due to its statistical nature, MI is sensitive to noise and artifacts and is not suitable for complex nonlinear deformations.

\textbf{Structural image representation}: Traditional structural image representation methods \cite{borvornvitchotikarn2020mirid,heinrich2012mind,jiang2016milbp,jaouen2021regularized} aim to extract spatially informative anatomical features that are more robust to intensity variations than raw pixel values. These approaches rely on hand-crafted descriptors designed to capture local structural patterns in a way that is less sensitive to differences in imaging protocols. While they do not explicitly model tissue deformations, such features are often more stable under small local changes. Registration algorithms then leverage these structural representations to estimate spatial transformations by comparing the similarity of the extracted features rather than relying directly on intensity values.

\textbf{Supervised metric learning}: Metric learning provides an alternative to traditional hand-crafted feature descriptors by leveraging deep neural networks to learn feature representations that minimize the distance between corresponding points in aligned image pairs \cite{simonovsky2016deep,blendowski2019combining}. This approach enables the extraction of discriminative features, tailored to capture complex anatomical variations and subtle tissue characteristics. However, supervised metric learning depends heavily on the availability of accurately aligned training data, which is often difficult and costly to obtain in medical imaging.

\textbf{Self-supervised contrastive learning in medical imaging}: Self-supervised contrastive learning has emerged as a powerful approach for representation learning in medical imaging, as it circumvents the need for explicit ground-truth annotations by enforcing consistency constraints on features extracted from augmented views of the same image \cite{wang2021dense,chaitanya2020contrastive,goncharov2023vox2vec,kats2024self,yan2022sam,bai2023samv2}. This paradigm enables models to learn robust, semantically meaningful representations by maximizing agreement between different transformations of the same sample, thereby promoting generalization across diverse, unlabeled datasets. Within this framework, data augmentation plays a central role in shaping the learned feature space. Intensity-based augmentations and the addition of various noise types encourage the model to focus on anatomical structural features that remain robust despite fluctuations in image appearance and quality. Similarly, geometric transformations, such as rotations, scaling, and elastic deformations, are critical for fostering invariance to spatial changes, ensuring that the learned global embeddings are resilient to differences in orientation, scale, and shape. Recent research has further advanced this direction by introducing equivariance constraints into the training objective \cite{pielawski2020comir,seince2024dense,santhirasekaram2024geometric}. Unlike invariance, which aims to ignore certain transformations, equivariance ensures that specific transformations in the input space lead to predictable and structured changes in the embedding space. This property allows the model to distinguish between meaningful anatomical variability and non-informative deformations.

\textbf{Contrastive learning in medical image registration}:
Recent studies have demonstrated the effectiveness of contrastive learning for generating dense feature representations that are well-suited to medical image registration tasks \cite{pielawski2020comir,li2023samconvex,liu2021same,mok2024modality,dey2022contrareg}. Existing methods in this space primarily differ in the stage at which feature extraction occurs relative to deformation estimation.

Some approaches perform feature extraction after the deformation field has been estimated. In this setting, features are computed from the fixed image and the deformed moving image, and the training objective is applied to corresponding feature representations. For instance, Mok et al. \cite{mok2024modality} pretrain a feature extractor using contrastive learning to enhance discriminability of anatomical structures across varying intensity profiles. During registration, the pretrained encoder is used to extract features from both fixed and warped moving images, and a Mean Square Error (MSE) loss is applied between them. ContraReg \cite{dey2022contrareg} leverages a contrastive objective to train a registration model. It applies the contrastive loss to dense, multi-scale feature maps extracted from fixed and warped images using a pretrained autoencoder. In both approaches, the feature extractors are pretrained independently and kept fixed during the registration training phase.

Other methods extract features prior to deformation and use them as inputs to the registration model. CoMIR \cite{pielawski2020comir} employs supervised contrastive learning on aligned multimodal image pairs to project them into a shared latent space. These learned representations are then used to train a monomodal registration network in a separate stage. SAMConvex \cite{li2023samconvex} and SAME \cite{liu2021same} incorporate Self-supervised Anatomical eMbeddings (SAM) \cite{yan2022sam} to improve registration performance. SAMConvex combines SAM-based embeddings with decoupled convex optimization strategies \cite{siebert2021fast,heinrich2014non}, while SAME utilizes SAM to estimate affine transformations, produce coarse deformation fields, and integrates these estimates into the VoxelMorph framework \cite{balakrishnan2019voxelmorph}.

\textbf{Contributions}: This paper introduces CoRe  - \textbf{Co}ntrastive learning for medical image \textbf{Re}gistration\footnote{The code is available at \url{https://github.com/EytanKats/reg-ssl}.}, a novel approach that integrates self-supervised contrastive learning with deformable image registration through a unified, jointly optimized framework. By aligning feature learning with the downstream registration task, CoRe produces dense, semantically meaningful representations tailored for accurate deformation field estimation. CoRe follows the pre-deformation feature extraction paradigm, where features are extracted independently from fixed and moving images prior to deformation estimation. These features are then used by a differentiable optimization module to infer the deformation field.

Unlike existing approaches such as CoMIR \cite{pielawski2020comir}, CoRe does not rely on known correspondences or paired multimodal training data. Instead, it leverages geometric augmentations of the same image to provide a self-supervised signal for contrastive learning. In contrast to SAMConvex \cite{li2023samconvex}, CoRe does not require a separate pretraining stage for the feature extractor. More importantly, it integrates knowledge of the registration task directly into the feature learning process via joint optimization, thereby aligning the learned representations with the requirements of the downstream registration module. As a result, the learned embeddings (1) capture semantic anatomical information that remains robust under tissue deformations and intensity variations, and (2) are inherently well-suited for driving the registration process through the optimization module.

The primary contributions of this work are as follows:
\begin{itemize}[itemsep=1pt,parsep=0pt,topsep=1pt,partopsep=1pt]
\item We propose a novel joint optimization framework that integrates self-supervised contrastive learning directly into the registration pipeline, enhancing feature robustness to anatomical variability and improving overall registration performance.
\item We demonstrate that simultaneous optimization of the contrastive and registration objectives leads to superior results compared to training with either objective alone, highlighting the synergy between representation learning and registration.
\item We validate CoRe on abdominal and thoracic datasets in both intra-patient and inter-patient settings, where it achieves superior performance compared to competitive approaches.
\end{itemize}  
\section{Materials and methods}
\label{sec:method}

\begin{figure*}[t]
   \includegraphics[width=1.0\linewidth]{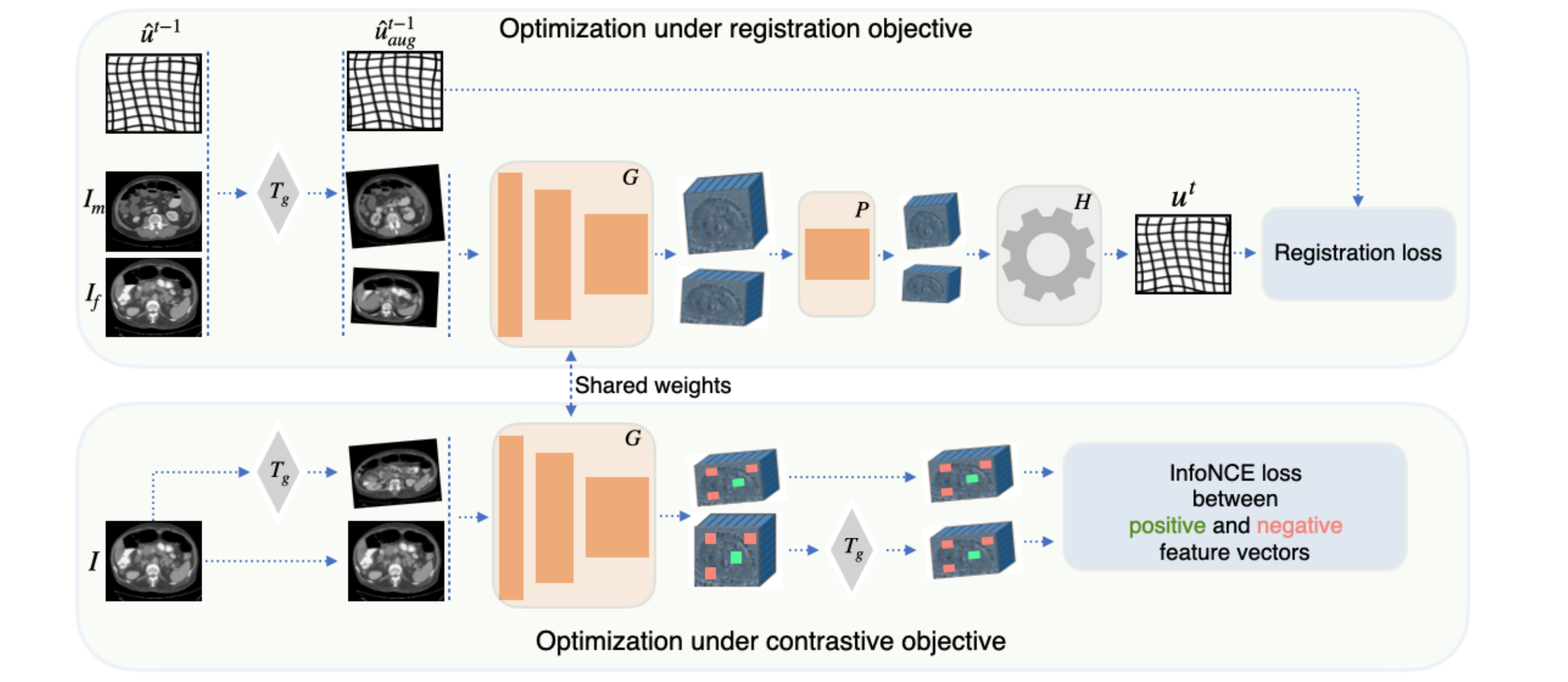}

   \caption{Overview of the proposed CoRe framework: The feature extractor is jointly optimized using registration and equivariance-based contrastive objectives, enabling robust and spatially coherent feature representations for precise displacement field estimation.}
   \label{fig:method}
\end{figure*}

\subsection{Problem definition}
\label{subsec:problem_definition}
Let $I_f, I_m$ denote the fixed and moving images, respectively. The training dataset consists of $|\Omega|$ image pairs $\Omega=\{I_f^s, I_m^s\}_{s=1}^{|\Omega|}$. The registration framework $\mathcal{R}$ comprises a trainable feature extractor $\mathcal{G}$ and a deterministic optimization module $\mathcal{H}$. Given $I_f$ and $I_m$ it predicts a displacement field $u = \mathcal{R}(I_f, I_m)$. Ideally, the intensity values $I_f(p)$ and $[Su \circ I_m](p)$ should correspond to the same anatomical location, where $[Su \circ I_m]$ represents $I_m$ warped by the spatial transformation $Su$ induced by $u$. The objective is to train $\mathcal{G}$ to extract high-quality features, enabling $\mathcal{H}$ to compute an optimal displacement field for accurate image alignment.

In this work, we incorporate equivariance constraints, formulated through a contrastive objective (Sec. \ref{subsec:constraint}), directly into the registration framework (Sec. \ref{subsec:reg_framework}). This integration ensures that the internal feature representations corresponding to identical anatomical locations remain robust to tissue deformations. Figure~\ref{fig:method} presents an overview of the proposed joint optimization strategy, highlighting the simultaneous optimization of the feature extractor under both contrastive and registration objectives. Alg. \ref{alg:im_features} outlines the pseudo-code for the training procedure, detailing the steps involved in leveraging the synergistic interaction between these two objectives to enhance registration accuracy and robustness.

\subsection{Registration framework}
\label{subsec:reg_framework}

We use a hybrid registration pipeline comprising a convolutional feature extractor $\mathcal{G}$, a convolutional projection head $\mathcal{P}$, and a differentiable optimization module $\mathcal{H}$, which, given fixed and moving features, infers a displacement field that minimizes a combined objective of smoothness and feature dissimilarity. To achieve this, optimization module $\mathcal{H}$ computes feature correlations over a discretized search window in a manner that remains differentiable with respect to the features \cite{siebert2022learn}. The pipeline begins with $\mathcal{G}$ extracting feature representations, $\mathcal{G}(I_f)$ and $\mathcal{G}(I_m)$, from the fixed and moving images, respectively. These representations are then passed through the projection head $\mathcal{P}$, and the resulting embeddings are processed by the optimizer $\mathcal{H}$, which predicts the displacement field $u$ as follows:

\begin{equation}
u = \mathcal{R}(I_f, I_m) = \mathcal{H}(\mathcal{P}(\mathcal{G}(I_f)), \mathcal{P}(\mathcal{G}(I_m))).
\end{equation}

We adopt the self-training scheme with pseudo-labels \cite{bigalke2023unsupervised} as a strong baseline for for deformable image registration. Training proceeds in $M$ stages. At the beginning of each stage $t = 1, \ldots, M$, the registration pipeline $\mathcal{R}$ generates displacement fields $u^{t-1}$ for all image pairs. These fields are refined through an iterative instance optimization, which includes (1) enforcing forward-backward consistency on the displacement field, (2) applying a double warping procedure, and (3) performing iterative instance optimization. The refined displacement fields $\hat{u}^{t-1}$ serve as pseudo-ground truth for supervising training of $\mathcal{G}$ and $\mathcal{P}$ during stage $t$. At the start of training, pseudo-labels $\hat{u}^{0}$ are generated using randomly initialized $\mathcal{G}$ and $\mathcal{P}$. The training objective minimizes the mean squared error (MSE) loss between the displacement fields $u^t$ predicted at training step of stage $t$ and the pseudo-labels $\hat{u}^{t-1}$ generated at the beginning of that stage:

\begin{equation}
\mathcal{L}_{\text{reg}} = \| u^t - {\hat{u}^{t-1}} \|^2.
\end{equation}

To enhance the diversity of transformations during training, augmentation is applied to the pseudo displacement field (Figure~\ref{fig:method}). Specifically, the fixed and moving images, $I_f$ and $I_m$, are each transformed using unique random affine augmentations $T_g^f$ and $T_g^m$, respectively. The pseudo displacement field ${\hat{u}}$ is then adjusted to account for these affine transformations, resulting in the augmented displacement field ${\hat{u}}_{aug}$.

\subsection{Equivariance constraint}
\label{subsec:constraint}

The quality of the displacement field $u$ generated by the optimizer $\mathcal{H}$ relies on the quality of the features extracted by the network $\mathcal{G}$. Ideally, the embeddings produced by $\mathcal{G}$ for the same anatomical location in the moving image $I_m$ and the fixed image $I_f$ should be identical, regardless of geometric deformations between the images. Such consistency in feature embeddings provides $\mathcal{H}$ with a robust initialization, allowing it to generate accurate displacement fields.

While the registration loss $\mathcal{L}_{\text{reg}}$ (Sec. \ref{subsec:reg_framework}) naturally improves the features extracted by $\mathcal{G}$ during training, we propose incorporating a contrastive objective to further refine feature quality. Specifically, to address the challenges posed by geometric deformations in tissue, we introduce an equivariance constraint on the image embeddings.

This constraint enforces consistency between embeddings derived from the same image $I$ under geometric transformations. We apply an affine transformation $T_g \sim \mathcal{T}_g$, sampled from a predefined augmentation set $\mathcal{T}_g$, to the image $I$. The constraint ensures consistency between the transformed features of the original image, $G_A = T_g(\mathcal{G}(I))$, and the features of the transformed image, $G_B = \mathcal{G}(T_g(I))$. By enforcing this property, the model learns representations that are resilient to tissue deformations - an essential requirement for registration tasks, where features must remain consistent across anatomical distortions.

We implement the equivariance constraint using an InfoNCE loss \cite{chen2020simple} applied to feature vectors extracted from corresponding spatial locations in the feature maps $G_A$ and $G_B$. Let $f_A^j$ and $f_B^j$ denote feature vectors sampled from the $j$th spatial location in $G_A$ and $G_B$, where $j = 1, \ldots, n$. Each feature vector $f_A^j$ forms one positive pair with the corresponding vector $f_B^j$ and $2 \cdot (n-1)$ negative pairs with other feature vectors sampled from $G_A$ and $G_B$. The contrastive loss is then defined as:

\begin{equation}
L_c = -\sum_j\log\frac{d(f_A^j, f_B^j)}{d(f_A^j, f_B^j) + \sum_{l \neq j}\sum_{k \in {A,B}}d(f_A^j, f_k^l)},
\end{equation}
where $d(f_A^j, f_B^j) = \exp\left(\langle f_A^j, f_B^j \rangle / \tau\right)$, $\langle \cdot, \cdot \rangle$ denotes the inner product, and $\tau$ is a temperature scaling factor that controls the sharpness of the similarity distribution. In our experiments, we set $\tau = 0.1$.

\begin{lstlisting}[style=mdpipython, caption={Pseudocode of the training procedure in PyTorch-like style.}, label={alg:im_features}, float, frame=tb]
for I_f, I_m, disp_gt in data_loader:

    # apply random affine transform
    I_f_aug, I_m_aug, disp_aug, T_f, T_m = aug_affine(I_f, I_m, disp_gt)
    
    # extract features from not augmented images
    feat_f = feature_extractor(I_f)
    feat_m = feature_extractor(I_m)
    
    # extract features from augmented images
    feat_f_aug_A = feature_extractor(I_f_aug)
    feat_m_aug_A = feature_extractor(I_m_aug)

    # extract projections from augmented features
    proj_f = projector(feat_f_aug_A)
    proj_m = projector(feat_m_aug_A)

    # generate displacement field
    disp = get_disp(proj_f, proj_m)
    
    # calculate registration loss
    L_reg = MSE(disp_aug, disp_gt)

    # transform features of not augmented images
    feat_f_aug_B = transform_feat(T_aff_f, feat_f)
    feat_m_aug_B = transform_feat(T_aff_m, feat_m)

    # calculate contrastive loss
    L_cl_f = InfoNCE(feat_f_aug_A, feat_f_aug_B)
    L_cl_m = InfoNCE(feat_m_aug_A, feat_m_aug_B)
    
    # calculate total loss
    L_total = L_reg + alpha * (L_cl_f + L_cl_m)

    # update weights
    loss.backward()
    optimizer.update([encoder, projector])
\end{lstlisting}

\subsection{Joint optimization}
\label{subsec:joint optimization}
By jointly minimizing the registration loss (Sec.\ref{subsec:reg_framework}) and the contrastive loss (Sec.\ref{subsec:constraint}), the proposed framework ensures that the feature representations extracted by the network $\mathcal{G}$ are robust to geometric transformations while remaining well-suited to the optimization procedure defined by the optimizer $\mathcal{H}$. The registration loss drives the alignment of fixed and moving images, encouraging the feature extractor to generate embeddings that are specifically tailored for consumption by the optimizer. Concurrently, the contrastive loss provides valuable guidance to the optimization process by imposing equivariance to geometric distortions, fostering the consistency of feature embeddings for same anatomical locations in registered images. The joint optimization process integrates the strengths of both losses, improving the robustness and accuracy of the registration framework. The combined loss function is defined as:

\begin{equation}
L = L_{reg} + \alpha \cdot L_c,
\label{eq:total_loss}
\end{equation}
where $\alpha$ is a weighting coefficient that balances the contributions of the contrastive loss and the registration objective.

\subsection{Implementation details}
\label{subsec:implementation_details}
The feature extractor $\mathcal{G}$ is composed of four convolutional blocks, where each block includes a 3D convolutional layer with a kernel size of $3\times3\times3$ followed by a batch normalization layer and a ReLU activation function. The projection head $\mathcal{P}$ consists of a single convolutional block with 128 output channels, a kernel size of $3\times3\times3$, and a stride of 2, followed by a final convolutional layer with a kernel size of $1\times1\times1$, which projects the feature maps to 16 channels. The framework is trained for $M = 8$ stages, with each stage consisting of 1000 iterations and a batch size of 2. Optimization is performed using the Adam optimizer, and the learning rate follows a cosine annealing warm restart schedule, decaying from $1\cdot10^{-3}$ to $1\cdot10^{-5}$. The contrastive loss is applied to the output of the final block's convolutional layer, with 1,000 feature vectors sampled per image pair.

\subsection{Datasets}
\label{subsec:datasets}
We evaluate the performance of the proposed method on the challenging inter-patient abdominal CT registration dataset \cite{xu2016evaluation}. This dataset comprises 30 3D abdominal CT scans from different patients, with 13 manually labeled anatomical structures: spleen, right kidney, left kidney, gall bladder, esophagus, liver, stomach, aorta, inferior vena cava, portal and splenic vein, pancreas, left adrenal gland, and right adrenal gland. All images are resampled to a uniform voxel resolution of 2 mm and standardized to spatial dimensions of $192 \times 160 \times 256$ voxels. The training-test split of this dataset defined in Learn2Reg challenge \cite{hering2022learn2reg} widely adapted in the medical image registration community which facilitates direct comparison with prior works. Specifically, the training set includes 20 scans (190 image pairs), while the test set consists of 10 scans (45 image pairs).

To evaluate performance in the intra-patient setting, we utilize the RAD-ChestCT dataset \cite{draelos2021machine}. In this dataset, we identified 371 longitudinal scan pairs. We split the data to 300 pairs designated for training and 71 pairs for testing. The CT images are resampled to a consistent voxel resolution of 1.5 mm and spatial dimensions of $256 \times 256 \times 224$ voxels. Since the RAD-ChestCT dataset does not include manual segmentation labels, we employ the TotalSegmentator tool \cite{wasserthal2023totalsegmentator} to segment the CT scans. Using the resulting segmentations, we calculate registration accuracy across 22 anatomical structures: 5 lung lobes, vertebrae from T1 to T12, heart myocardium, left and right heart ventricles and atriums.
\section{Results and discussion}
\label{sec:experiments}

\begin{table*}[t]
    \caption{Quantitative results for AbdomenCT and RadChestCT registration tasks. The proposed CoRe method outperforms all comparison methods in terms of Dice, with statistical significance confirmed by a Wilcoxon signed-rank test (p < 0.0001), underscoring the advantages of the joint optimization strategy. The smoothness of the predicted displacement fields (SDlogJ) remains low and comparable to other methods.}
     \begin{adjustwidth}{-\extralength}{0cm}
		\begin{tabularx}{\fulllength}{cCCCCCC}
			\toprule[1.5pt]
			\multirow{2}{*}{Method} & \multicolumn{3}{c}{AbdomenCT} & \multicolumn{3}{c}{RadChestCT}\\
                \cmidrule(lr){2-4}\cmidrule(lr){5-7}
			 & \rule{1pt}{0ex} $\text{DSC}\uparrow$ & $\text{SDLogJ}\downarrow$ & $\text{T}_{inf}\downarrow$ & \rule{1pt}{0ex} $\text{DSC}\uparrow$ & \rule{1pt}{0ex}  $\text{SDLogJ}\downarrow$ & $\text{T}_{inf}\downarrow$ \\
			\midrule[1pt]
			Initial \hspace{0.1cm} & 25.9 $\pm$ 7.1 & -- & --  & 34.1 $\pm$ 14.5 & -- & -- \\
                \midrule
                DEEDs~\cite{heinrich2013mrf} \hspace{0.1cm} & 46.5 $\pm$ 8.1 & 0.058 & 45.2 s  & 88.5 $\pm$ 6.9 & 0.049 & 45.0 s \\
                NiftyReg~\cite{modat2010fast} \hspace{0.1cm} & 34.9 $\pm$ 9.3 & \textbf{0.034} & 123.5 s  & 84.4 $\pm$ 7.4 & \textbf{0.021} & 85.5 s \\
                \midrule
                VoxelMorph~\cite{balakrishnan2019voxelmorph} \hspace{0.1cm} & 35.4 $\pm$ 8.9 & 0.134 & \textbf{0.2 s}  & 55.6 $\pm$ 10.3 &  0.101 & \textbf{0.3 s} \\
                LapIRN~\cite{mok2020large} \hspace{0.1cm} & 42.4 $\pm$ 8.2 & 0.089 & 0.7 s & 83.4 $\pm$ 7.6 &  0.075 & 0.8 s \\
                uniGradICON~\cite{tian2024unigradicon} \hspace{0.1cm} & 52.1 $\pm$ 6.9  & 0.117 & 39.7 s  & 86.8 $\pm$ 8.3 &  0.092 & 45.6 s \\
                \midrule
                SAMConvex~\cite{li2023samconvex} \hspace{0.1cm} & 51.2 $\pm$ 7.8  & 0.096 & 5.1 s  & 88.1 $\pm$ 7.1 &  0.079 & 6.4 s \\
                Bigalke~et~al.~\cite{bigalke2023unsupervised} \hspace{0.1cm} & 51.1 $\pm$ 7.3 & 0.146 & 1.2 s & 87.3 $\pm$ 6.9 &  0.106 & 1.9 s \\
                CoRe (ours) \hspace{0.1cm} & \hspace{0.1cm} \textbf{52.6 $\pm$ 7.5} \hspace{0.1cm} & \hspace{0.1cm} 0.148 \hspace{0.1cm} & \hspace{0.1cm} 1.2 s \hspace{0.1cm} & \hspace{0.1cm} \textbf{89.4 $\pm$ 6.6} \hspace{0.1cm} & \hspace{0.1cm} 0.109 \hspace{0.1cm} & \hspace{0.1cm} 1.9 s \hspace{0.1cm} \\
			\bottomrule[1.5pt]
		\end{tabularx}
    \end{adjustwidth}
	\label{tab:main_result}
\end{table*}

To assess accuracy of the registration, we compute the average Dice similarity coefficient ($DSC$) using available segmented structures. The plausibility of the deformation fields is evaluated using the standard deviation of the logarithm of the Jacobian determinant ($SDlogJ$). Additionally, we report inference run-time ($T_{inf}$) across methods.

\subsection{Registration results}
\label{subsec:registration_results}

We compare our method with conventional registration approaches (NiftyReg \cite{modat2010fast} and DEEDs \cite{heinrich2013mrf}), learning-based methods (VoxelMorph \cite{balakrishnan2019voxelmorph}, LapIRN \cite{mok2020large} and uniGradIcon \cite{tian2024unigradicon}), and two hybrid approaches (Bigalke et~al. \cite{bigalke2023unsupervised} and SAMConvex \cite{li2023samconvex}) (Table~\ref{tab:main_result}). NiftyReg utilizes a four-level multi-resolution strategy, adaptive gradient descent optimization, and mutual information (MI) as the similarity measure. DEEDs aligns images using edge-based similarity and employs a B-spline transformation model with diffusion regularization. VoxelMorph and LapIRN predict dense displacement fields directly from image pairs using convolutional neural networks, with LapIRN refining the deformation across multiple resolution levels. uniGradICON builds upon the gradient inverse consistency (GradICON \cite{tian2023gradicon}) principle to improve robustness across datasets and is trained on a diverse collection of data. During inference, we employ the instance-specific optimization option provided by uniGradICON, which fine-tunes the pretrained model weights for each image pair to achieve improved performance. Bigalke et~al. and SAMConvex are hybrid approaches that leverage CNNs for feature extraction from image pairs and classical optimization techniques for displacement field estimation. SAMConvex uses a pretrained SAM model \cite{yan2022sam} for feature extraction, while Bigalke et~al. optimize the feature extractor with a differentiable optimizer and registration loss (Figure~\ref{fig:highlight}).

\begin{figure*}[t]
   \includegraphics[width=1\linewidth]{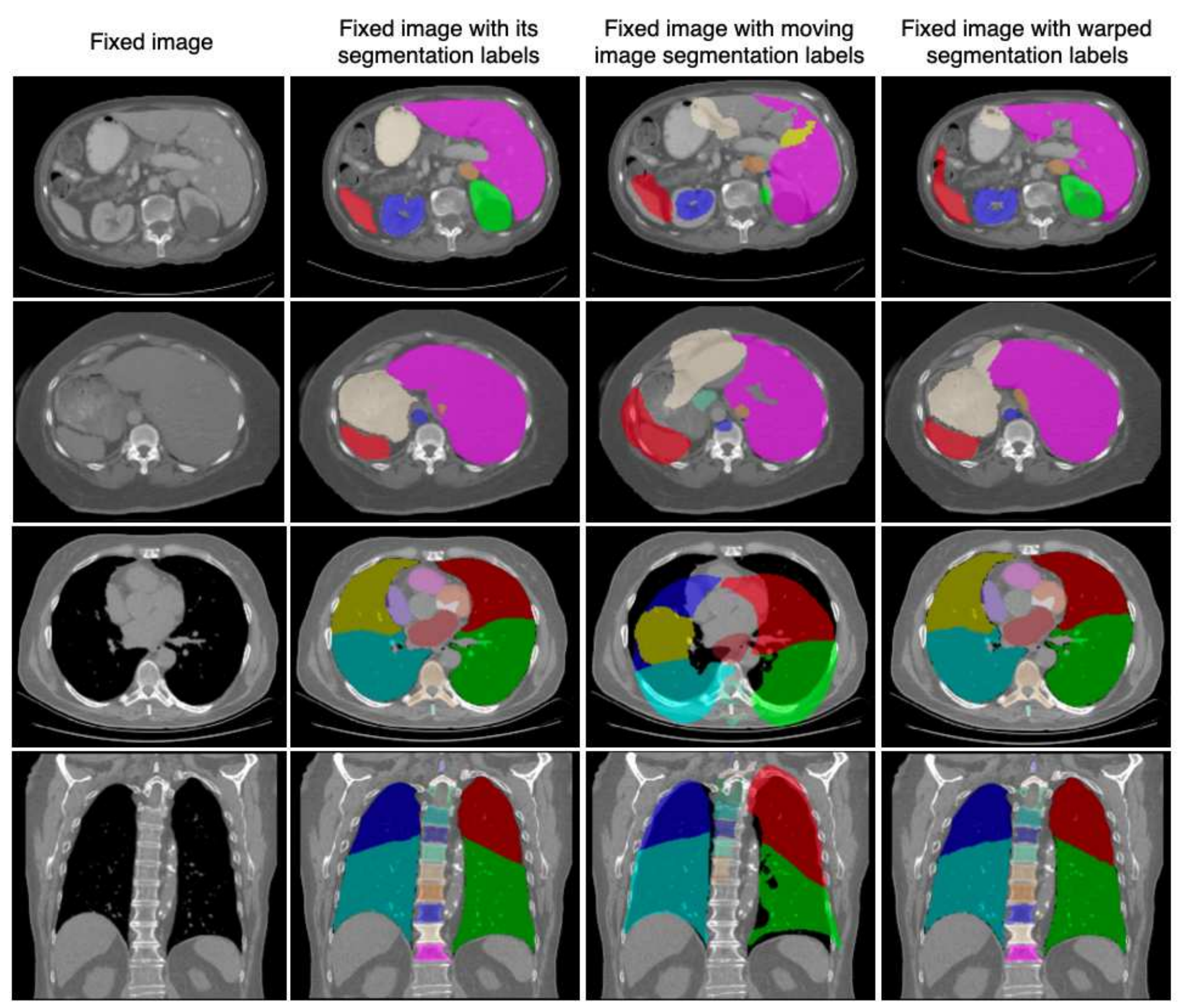}

   \caption{Qualitative results of the proposed CoRe method. From left to right: fixed image, fixed image with its segmentation overlay, fixed image with the overlay of the moving image segmentation, and fixed image with the overlay of the warped segmentation. The top two rows show examples from the AbdomenCT dataset in the axial plane, while the bottom two rows present examples from the RadChestCT dataset in axial and coronal planes.}
   \label{fig:methods_comparison_abdomen}
\end{figure*}

VoxelMorph and LapIRN are computationally efficient, however, they often underperform compared to traditional and hybrid methods. uniGradICON achieves strong results on both datasets, but relies on instance-specific optimization during inference, which leads to longer inference times. DEEDS achieves strong results on the RadChestCT dataset, ranking as the second-best method. This performance is expected due to its focus on optimizing edge similarity, which is highly effective for intra-patient thoracic datasets where edges in image pairs align closely. However, on the AbdomenCT dataset, where deformations between image pairs are more complex, DEEDS demonstrates lower accuracy compared to hybrid methods. Hybrid approaches combine deep learning's ability to extract robust features with the precision and reliability of classical optimization techniques for displacement field estimation. This synergy enables hybrid methods to achieve state-of-the-art performance on the challenging inter-patient AbdomenCT dataset while maintaining competitive results on RadChestCT. Our proposed CoRe method achieves the best performance on both datasets, delivering the highest Dice scores (DSC) while preserving smoothness in the predicted displacement fields (SDLogJ), comparable to competitive methods. These results underscore the effectiveness of our approach, which incorporates an equivariance-based contrastive objective directly into the registration framework, enabling performance improvement for image registration tasks. Figure~\ref{fig:methods_comparison_abdomen} presents qualitative registration results of the proposed method on the AbdomenCT and RadChestCT datasets.

\subsection{Ablations study}
\label{subsec:ablations}

To assess the effectiveness of the proposed method, we trained the feature extractor $\mathcal{G}$ using regularization loss and contrastive loss independently and compared the results with the proposed joint optimization approach (Table~\ref{tab:ablation_loss}). For the contrastive loss, we initially pretrained $\mathcal{G}$ using only contrastive objective and subsequently trained the registration framework with $\mathcal{G}$ frozen using only registration objective. The joint optimization approach demonstrates superior performance, underscoring the synergistic benefits of combining these objectives. This strategy facilitates the extraction of more discriminative and spatially coherent features, enhancing registration accuracy across datasets.

\begin{table}[t]
    \caption{Comparison of training the feature extractor with registration and contrastive objectives separately and jointly. The proposed joint optimization strategy achieves the best performance.
    }
		\begin{tabular}{cccccc}
			\toprule[1.5pt]
			\multirow{2}{*}{$\mathcal{L}_{reg}$} & \multirow{2}{*}{$\mathcal{L}_{cl}$} & \multicolumn{2}{c}{AbdomenCT} & \multicolumn{2}{c}{RadChestCT}\\
                \cmidrule(lr){3-4}\cmidrule(lr){5-6}
			& & \rule{1pt}{0ex} $\text{DSC}\uparrow$ & $\text{SDLogJ}\downarrow$ & \rule{1pt}{0ex} $\text{DSC}\uparrow$ & \rule{1pt}{0ex}  $\text{SDLogJ}\downarrow$ \\
			\midrule[1pt]
                \checkmark \hspace{0.1cm} & & 51.1 $\pm$ 7.3 & \textbf{0.146} & 87.3 $\pm$ 6.9 &  \textbf{0.106} \\
                & \checkmark & 50.3 $\pm$ 7.5 & 0.151 & 86.7 $\pm$ 7.1 &  0.117 \\
                \checkmark & \checkmark & \hspace{0.1cm} \textbf{52.6 $\pm$ 7.5} \hspace{0.1cm} & \hspace{0.1cm} 0.148 \hspace{0.1cm} & \hspace{0.1cm} \textbf{89.4 $\pm$ 6.6} \hspace{0.1cm} & \hspace{0.1cm} 0.109 \hspace{0.1cm}\\
			\bottomrule[1.5pt]
		\end{tabular}
    \label{tab:ablation_loss}
\end{table}

\begin{table}{H}
    \caption{Comparison of the effect of intensity invariance and geometric equivariance constraints, applied separately and jointly. Training with the equivariance constraint achieves the best performance, highlighting the importance of feature robustness to non-linear tissue deformations.
    }
		\begin{tabular}{cccccc}
			\toprule[1.5pt]
			\multirow{2}{*}{$T_{i}$} & \multirow{2}{*}{$T_{g}$} & \multicolumn{2}{c}{AbdomenCT} & \multicolumn{2}{c}{RadChestCT}\\
                \cmidrule(lr){3-4}\cmidrule(lr){5-6}
			& & \rule{1pt}{0ex} $\text{DSC}\uparrow$ & $\text{SDLogJ}\downarrow$ & \rule{1pt}{0ex} $\text{DSC}\uparrow$ & \rule{1pt}{0ex}  $\text{SDLogJ}\downarrow$ \\
			\midrule[1pt]
                & & 51.1 $\pm$ 7.3 & 0.146 & 87.3 $\pm$ 6.9 &  0.106 \\
                \checkmark \hspace{0.1cm} & & 49.4 $\pm$ 8.1  & 0.139 & 87.5 $\pm$ 7.0 &  0.115 \\
                & \checkmark & \textbf{52.6 $\pm$ 7.5} & 0.148 & \textbf{89.4 $\pm$ 6.6} &  0.109 \\
                \checkmark & \checkmark & \hspace{0.1cm} 50.2 $\pm$ 7.6 \hspace{0.1cm} & \hspace{0.1cm} \textbf{0.138} \hspace{0.1cm} & \hspace{0.1cm} 89.3 $\pm$ 6.5 \hspace{0.1cm} & \hspace{0.1cm} \textbf{0.102} \hspace{0.1cm}\\
			\bottomrule[1.5pt]
		\end{tabular}
    \label{tab:ablation_augmentations}
    
\end{table}

Along with the equivariance constraint described in Sec. \ref{subsec:constraint}, self-supervised contrastive learning methods commonly employ non-linear intensity augmentations during pretraining to encourage feature invariance to variations in appearance while preserving spatial encoding. To evaluate the impact of these augmentations on the registration task, we trained our framework with geometric equivariance and appearance invariance constraints both independently and jointly. Interestingly, the results reveal that within the proposed framework, non-linear intensity augmentations do not provide additional benefits over training solely with the geometric equivariance constraint. For the AbdomenCT dataset, training with intensity augmentations for contrastive loss even results in inferior performance compared to using the registration objective alone. We hypothesize that this is due to the mono-modal nature of the CT datasets used in our evaluation. The standardized intensity values in the CT datasets may limit the effectiveness of intensity augmentations, as they do not enhance the discriminative capacity of the learned features. Future work may explore the utility of intensity augmentations in multi-modal settings or datasets with greater intensity variability, where these augmentations could play a more significant role in improving registration performance.

\begin{figure}
\subfloat[\centering]{\includegraphics[width=7.0cm, height=5.0cm]{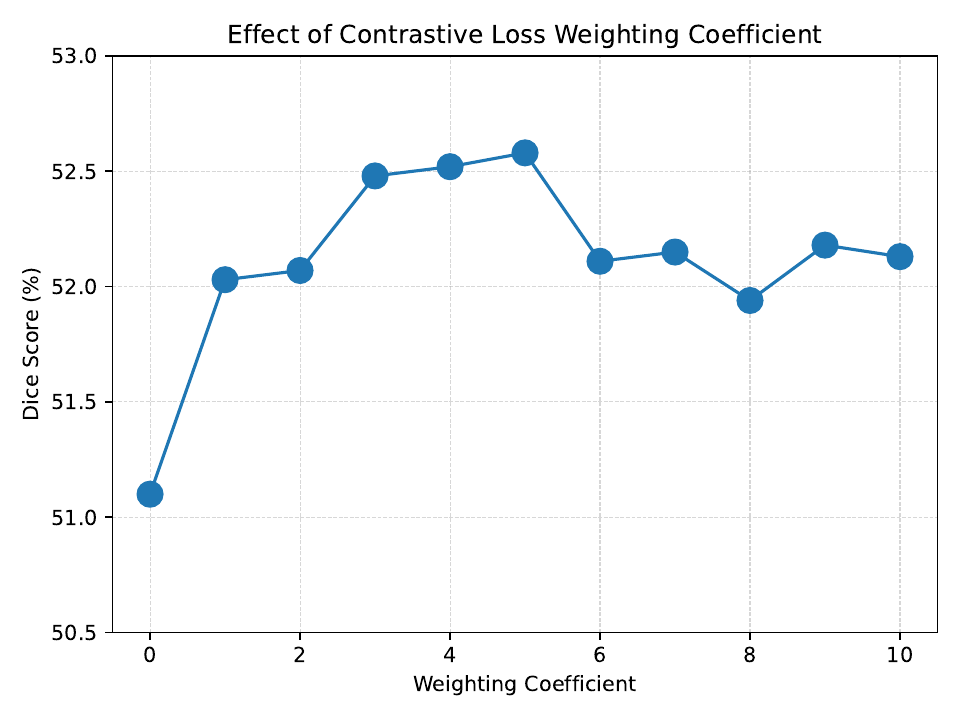}
\label{fig:lambda_plot}}
\subfloat[\centering]{\includegraphics[width=7.0cm, height=5.0cm]{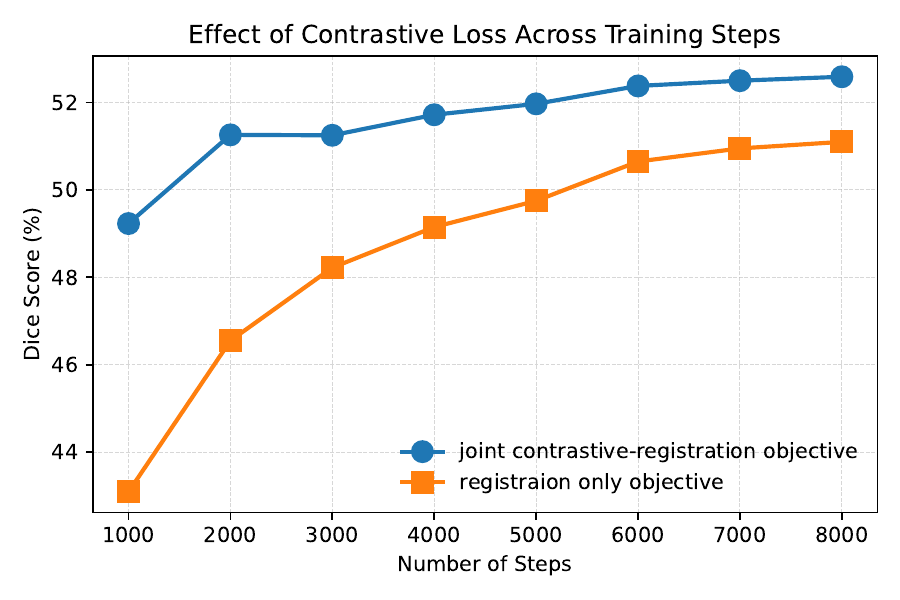}
\label{fig:iterations_plot}}\\
\caption{(a) Influence of the contrastive loss weighting coefficient $\alpha$ on registration performance. (b) Evolution of the Dice score over training iterations, comparing the proposed joint optimization strategy with a baseline trained using only the registration loss.}
\end{figure}

We further evaluate the performance of the proposed joint contrastive-registration framework with respect to different values of the weighting coefficient $\alpha$ (Figure~\ref{fig:lambda_plot}), which controls the contribution of the contrastive loss in the total objective (Equation~\ref{eq:total_loss}). Incorporating the contrastive component with $\alpha = 1$ already yields a measurable improvement, increasing the Dice score by 0.93\% compared to the baseline trained without contrastive supervision. As $\alpha$ increases, we observe a gradual improvement in performance, suggesting that a stronger emphasis on the contrastive objective encourages the learning of more robust and deformation-consistent feature representations. The best performance, with a Dice score of 52.59\%, is achieved at $\alpha = 5$, indicating a favorable balance between registration accuracy and representation learning. Increasing $\alpha$ beyond this value leads to a decline in performance, which may indicate that excessive weighting of the contrastive objective can interfere with the optimization of the registration task. Nevertheless, even at higher values of $\alpha$, the proposed framework consistently outperforms the baseline, supporting the effectiveness of the joint optimization strategy.

Figure~\ref{fig:iterations_plot} illustrates the effect of the contrastive loss across different stages of training. The most pronounced improvement over the baseline is observed during the early training phases, highlighting the impact of contrastive supervision in guiding the optimization process. The contrastive objective provides an informative learning signal at the beginning of training, enabling the model to converge more rapidly toward meaningful feature representations that are beneficial for registration. This is reflected in a performance gap of 6.12\% Dice after the first 1000 iterations. After only 2000 iterations, corresponding to one quarter of the total training, the proposed joint optimization strategy already achieves a Dice score of 51.26\%, surpassing the final performance of the baseline, 51.1\% Dice. Although the performance gap decreases as training progresses, it remains significant throughout the optimization and persists until convergence. These results suggest that integrating contrastive learning not only improves final performance but also contributes to faster convergence.

\section{Conclusions}
\label{sec:conclusions}

We introduced CoRe, a hybrid image registration framework that integrates contrastive learning into the registration pipeline. We demonstrated that jointly optimizing the feature extractor under both contrastive and registration objectives facilitates the learning of spatially coherent and discriminative features, tailored to the requirements of classical optimization procedures. Our findings emphasize the important role of equivariant geometric constraints, implemented through contrastive loss, in enabling the extraction of robust features. These features are particularly effective in handling tissue deformations, thereby improving registration performance. In addition, our analysis shows that the inclusion of contrastive supervision accelerates convergence, especially during the early stages of training, where the model benefits from a stronger and more informative learning signal. Quantitative results on the AbdomenCT and RadChestCT datasets confirm that CoRe consistently outperforms conventional, learning-based, and hybrid registration methods, achieving superior Dice scores while maintaining competitive smoothness of the displacement fields.

These results highlight the potential of directly integrating contrastive learning into the registration framework to improve medical image registration performance. By leveraging the synergies of joint optimization, CoRe provides a robust and accurate framework for efficient image alignment, while also enabling faster convergence during training. This makes it a promising direction for future research in medical image registration tasks.


\authorcontributions{Conceptualization, E.K. and M.H.; methodology, E.K.; software, E.K.; validation, E.K., C.G., Z.H., F.F, and W.H; formal analysis, E.K.; investigation, E.K.; resources, M.H.; data curation, E.K.; writing---original draft preparation, E.K.; writing---review and editing, E.K and M.H.; visualization, E.K.; supervision, M.H.; project administration, M.H.; funding acquisition, M.H. All authors have read and agreed to the published version of the manuscript.}

\funding{This research was funded by German Research Foundation: DFG, HE 7364/10-1, project number 500498869.}

\dataavailability{The datasets analyzed in this study are publicly accessible through third-party repositories. The inter-patient abdominal CT registration dataset [27] is available via Zenodo at \url{https://doi.org/10.5281/zenodo.3715652}. The RAD-ChestCT dataset is available via Zenodo at \url{https://doi.org/10.5281/zenodo.6406114}.}

\conflictsofinterest{The authors declare no conflicts of interest. The funders had no role in the design of the study; in the collection, analyses, or interpretation of data; in the writing of the manuscript; or in the decision to publish the results.} 

\begin{adjustwidth}{-\extralength}{0cm}

\reftitle{References}


\bibliography{main}

\begin{thebibliography}{999}

\bibitem[Li et~al.(2023)Li, Tian, Mok, Bai, Wang, Ge, Zhou, Lu, Ye, Yan, et~al.]{li2023samconvex}
Li, Z.; Tian, L.; Mok, T.C.; Bai, X.; Wang, P.; Ge, J.; Zhou, J.; Lu, L.; Ye, X.; Yan, K.;  et~al.
\newblock Samconvex: Fast discrete optimization for ct registration using self-supervised anatomical embedding and correlation pyramid.
\newblock In Proceedings of the International Conference on Medical Image Computing and Computer-Assisted Intervention. Springer,  2023, pp. 559--569.

\bibitem[Bigalke et~al.(2023)Bigalke, Hansen, Mok, and Heinrich]{bigalke2023unsupervised}
Bigalke, A.; Hansen, L.; Mok, T.C.; Heinrich, M.P.
\newblock Unsupervised 3d registration through optimization-guided cyclical self-training.
\newblock In Proceedings of the International Conference on Medical Image Computing and Computer-Assisted Intervention. Springer,  2023, pp. 677--687.

\bibitem[Maes et~al.(1997)Maes, Collignon, Vandermeulen, Marchal, and Suetens]{maes1997multimodality}
Maes, F.; Collignon, A.; Vandermeulen, D.; Marchal, G.; Suetens, P.
\newblock Multimodality image registration by maximization of mutual information.
\newblock {\em IEEE Transactions on Medical Imaging} {\bf 1997}, {\em 16},~187--198.

\bibitem[Borvornvitchotikarn and Kurutach(2020)]{borvornvitchotikarn2020mirid}
Borvornvitchotikarn, T.; Kurutach, W.
\newblock mirid: Multi-modal image registration using modality-independent and rotation-invariant descriptor.
\newblock {\em Symmetry} {\bf 2020}, {\em 12},~2078.

\bibitem[Heinrich et~al.(2012)Heinrich, Jenkinson, Bhushan, Matin, Gleeson, Brady, and Schnabel]{heinrich2012mind}
Heinrich, M.P.; Jenkinson, M.; Bhushan, M.; Matin, T.; Gleeson, F.V.; Brady, M.; Schnabel, J.A.
\newblock MIND: Modality independent neighbourhood descriptor for multi-modal deformable registration.
\newblock {\em Medical Image Analysis} {\bf 2012}, {\em 16},~1423--1435.

\bibitem[Jiang et~al.(2016)Jiang, Shi, Yao, Wang, and Song]{jiang2016milbp}
Jiang, D.; Shi, Y.; Yao, D.; Wang, M.; Song, Z.
\newblock miLBP: a robust and fast modality-independent 3D LBP for multimodal deformable registration.
\newblock {\em International journal of computer assisted radiology and surgery} {\bf 2016}, {\em 11},~997--1005.

\bibitem[Jaouen et~al.(2021)Jaouen, Conze, Dardenne, Bert, and Visvikis]{jaouen2021regularized}
Jaouen, V.; Conze, P.H.; Dardenne, G.; Bert, J.; Visvikis, D.
\newblock Regularized directional representations for medical image registration.
\newblock {\em arXiv preprint arXiv:2111.15509} {\bf 2021}.

\bibitem[Simonovsky et~al.(2016)Simonovsky, Guti{\'e}rrez-Becker, Mateus, Navab, and Komodakis]{simonovsky2016deep}
Simonovsky, M.; Guti{\'e}rrez-Becker, B.; Mateus, D.; Navab, N.; Komodakis, N.
\newblock A deep metric for multimodal registration.
\newblock In Proceedings of the Medical Image Computing and Computer-Assisted Intervention. Springer,  2016, pp. 10--18.

\bibitem[Blendowski and Heinrich(2019)]{blendowski2019combining}
Blendowski, M.; Heinrich, M.P.
\newblock Combining MRF-based deformable registration and deep binary 3D-CNN descriptors for large lung motion estimation in COPD patients.
\newblock {\em International journal of computer assisted radiology and surgery} {\bf 2019}, {\em 14},~43--52.

\bibitem[Wang et~al.(2021)Wang, Zhang, Shen, Kong, and Li]{wang2021dense}
Wang, X.; Zhang, R.; Shen, C.; Kong, T.; Li, L.
\newblock Dense contrastive learning for self-supervised visual pre-training.
\newblock In Proceedings of the Proceedings of the IEEE/CVF conference on computer vision and pattern recognition,  2021, pp. 3024--3033.

\bibitem[Chaitanya et~al.(2020)Chaitanya, Erdil, Karani, and Konukoglu]{chaitanya2020contrastive}
Chaitanya, K.; Erdil, E.; Karani, N.; Konukoglu, E.
\newblock Contrastive learning of global and local features for medical image segmentation with limited annotations.
\newblock {\em Advances in neural information processing systems} {\bf 2020}, {\em 33},~12546--12558.

\bibitem[Goncharov et~al.(2023)Goncharov, Soboleva, Kurmukov, Pisov, and Belyaev]{goncharov2023vox2vec}
Goncharov, M.; Soboleva, V.; Kurmukov, A.; Pisov, M.; Belyaev, M.
\newblock vox2vec: A framework for self-supervised contrastive learning of voxel-level representations in medical images.
\newblock In Proceedings of the International Conference on Medical Image Computing and Computer-Assisted Intervention. Springer,  2023, pp. 605--614.

\bibitem[Kats et~al.(2024)Kats, Hirsch, and Heinrich]{kats2024self}
Kats, E.; Hirsch, J.G.; Heinrich, M.P.
\newblock Self-supervised Learning of Dense Hierarchical Representations for Medical Image Segmentation.
\newblock {\em arXiv preprint arXiv:2401.06473} {\bf 2024}.

\bibitem[Yan et~al.(2022)Yan, Cai, Jin, Miao, Guo, Harrison, Tang, Xiao, Lu, and Lu]{yan2022sam}
Yan, K.; Cai, J.; Jin, D.; Miao, S.; Guo, D.; Harrison, A.P.; Tang, Y.; Xiao, J.; Lu, J.; Lu, L.
\newblock SAM: Self-supervised learning of pixel-wise anatomical embeddings in radiological images.
\newblock {\em IEEE Transactions on Medical Imaging} {\bf 2022}, {\em 41},~2658--2669.

\bibitem[Bai et~al.(2023)Bai, Bai, Huo, Ge, Lu, Ye, Yan, and Xia]{bai2023samv2}
Bai, X.; Bai, F.; Huo, X.; Ge, J.; Lu, J.; Ye, X.; Yan, K.; Xia, Y.
\newblock SAMv2: A Unified Framework for Learning Appearance, Semantic and Cross-Modality Anatomical Embeddings.
\newblock {\em arXiv preprint arXiv:2311.15111} {\bf 2023}.

\bibitem[Pielawski et~al.(2020)Pielawski, Wetzer, {\"O}fverstedt, Lu, W{\"a}hlby, Lindblad, and Sladoje]{pielawski2020comir}
Pielawski, N.; Wetzer, E.; {\"O}fverstedt, J.; Lu, J.; W{\"a}hlby, C.; Lindblad, J.; Sladoje, N.
\newblock CoMIR: Contrastive multimodal image representation for registration.
\newblock {\em Advances in neural information processing systems} {\bf 2020}, {\em 33},~18433--18444.

\bibitem[Seince et~al.(2024)Seince, Folgoc, de~Souza, and Angelini]{seince2024dense}
Seince, M.; Folgoc, L.L.; de~Souza, L.A.F.; Angelini, E.
\newblock Dense Self-Supervised Learning for Medical Image Segmentation.
\newblock {\em arXiv preprint arXiv:2407.20395} {\bf 2024}.

\bibitem[Santhirasekaram et~al.(2024)Santhirasekaram, Winkler, Rockall, and Glocker]{santhirasekaram2024geometric}
Santhirasekaram, A.; Winkler, M.; Rockall, A.; Glocker, B.
\newblock A geometric approach to robust medical image segmentation.
\newblock {\em Medical Image Analysis} {\bf 2024}, {\em 97},~103260.

\bibitem[Liu et~al.(2021)Liu, Yan, Harrison, Guo, Lu, Yuille, Huang, Xie, Xiao, Ye, et~al.]{liu2021same}
Liu, F.; Yan, K.; Harrison, A.P.; Guo, D.; Lu, L.; Yuille, A.L.; Huang, L.; Xie, G.; Xiao, J.; Ye, X.;  et~al.
\newblock SAME: Deformable image registration based on self-supervised anatomical embeddings.
\newblock In Proceedings of the International Conference on Medical Image Computing and Computer Assisted Intervention. Springer,  2021, pp. 87--97.

\bibitem[Mok et~al.(2024)Mok, Li, Bai, Zhang, Liu, Zhou, Yan, Jin, Shi, Yin, et~al.]{mok2024modality}
Mok, T.C.; Li, Z.; Bai, Y.; Zhang, J.; Liu, W.; Zhou, Y.J.; Yan, K.; Jin, D.; Shi, Y.; Yin, X.;  et~al.
\newblock Modality-Agnostic Structural Image Representation Learning for Deformable Multi-Modality Medical Image Registration.
\newblock In Proceedings of the IEEE/CVF Conference on Computer Vision and Pattern Recognition,  2024, pp. 11215--11225.

\bibitem[Dey et~al.(2022)Dey, Schlemper, Salehi, Zhou, Gerig, and Sofka]{dey2022contrareg}
Dey, N.; Schlemper, J.; Salehi, S.S.M.; Zhou, B.; Gerig, G.; Sofka, M.
\newblock Contrareg: Contrastive learning of multi-modality unsupervised deformable image registration.
\newblock In Proceedings of the International Conference on Medical Image Computing and Computer-Assisted Intervention. Springer,  2022, pp. 66--77.

\bibitem[Siebert et~al.(2021)Siebert, Hansen, and Heinrich]{siebert2021fast}
Siebert, H.; Hansen, L.; Heinrich, M.P.
\newblock Fast 3D registration with accurate optimisation and little learning for Learn2Reg 2021.
\newblock In Proceedings of the International Conference on Medical Image Computing and Computer-Assisted Intervention. Springer,  2021, pp. 174--179.

\bibitem[Heinrich et~al.(2014)Heinrich, Papie{\.z}, Schnabel, and Handels]{heinrich2014non}
Heinrich, M.P.; Papie{\.z}, B.W.; Schnabel, J.A.; Handels, H.
\newblock Non-parametric discrete registration with convex optimisation.
\newblock In Proceedings of the International Workshop on Biomedical Image Registration. Springer,  2014, pp. 51--61.

\bibitem[Balakrishnan et~al.(2019)Balakrishnan, Zhao, Sabuncu, Guttag, and Dalca]{balakrishnan2019voxelmorph}
Balakrishnan, G.; Zhao, A.; Sabuncu, M.R.; Guttag, J.; Dalca, A.V.
\newblock Voxelmorph: a learning framework for deformable medical image registration.
\newblock {\em IEEE Transactions on Medical Imaging} {\bf 2019}, {\em 38},~1788--1800.

\bibitem[Siebert and Heinrich(2022)]{siebert2022learn}
Siebert, H.; Heinrich, M.P.
\newblock Learn to fuse input features for large-deformation registration with differentiable convex-discrete optimisation.
\newblock In Proceedings of the International Workshop on Biomedical Image Registration. Springer,  2022, pp. 119--123.

\bibitem[Chen et~al.(2020)Chen, Kornblith, Norouzi, and Hinton]{chen2020simple}
Chen, T.; Kornblith, S.; Norouzi, M.; Hinton, G.
\newblock A simple framework for contrastive learning of visual representations.
\newblock In Proceedings of the International conference on machine learning. PMLR,  2020, pp. 1597--1607.

\bibitem[Xu et~al.(2016)Xu, Lee, Heinrich, Modat, Rueckert, Ourselin, Abramson, and Landman]{xu2016evaluation}
Xu, Z.; Lee, C.P.; Heinrich, M.P.; Modat, M.; Rueckert, D.; Ourselin, S.; Abramson, R.G.; Landman, B.A.
\newblock Evaluation of six registration methods for the human abdomen on clinically acquired CT.
\newblock {\em IEEE Transactions on Biomedical Engineering} {\bf 2016}, {\em 63},~1563--1572.

\bibitem[Hering et~al.(2022)Hering, Hansen, Mok, Chung, Siebert, H{\"a}ger, Lange, Kuckertz, Heldmann, Shao, et~al.]{hering2022learn2reg}
Hering, A.; Hansen, L.; Mok, T.C.; Chung, A.C.; Siebert, H.; H{\"a}ger, S.; Lange, A.; Kuckertz, S.; Heldmann, S.; Shao, W.;  et~al.
\newblock Learn2Reg: comprehensive multi-task medical image registration challenge, dataset and evaluation in the era of deep learning.
\newblock {\em IEEE Transactions on Medical Imaging} {\bf 2022}, {\em 42},~697--712.

\bibitem[Draelos et~al.(2021)Draelos, Dov, Mazurowski, Lo, Henao, Rubin, and Carin]{draelos2021machine}
Draelos, R.L.; Dov, D.; Mazurowski, M.A.; Lo, J.Y.; Henao, R.; Rubin, G.D.; Carin, L.
\newblock Machine-learning-based multiple abnormality prediction with large-scale chest computed tomography volumes.
\newblock {\em Medical image analysis} {\bf 2021}, {\em 67},~101857.

\bibitem[Wasserthal et~al.(2023)Wasserthal, Breit, Meyer, Pradella, Hinck, Sauter, Heye, Boll, Cyriac, Yang, et~al.]{wasserthal2023totalsegmentator}
Wasserthal, J.; Breit, H.C.; Meyer, M.T.; Pradella, M.; Hinck, D.; Sauter, A.W.; Heye, T.; Boll, D.T.; Cyriac, J.; Yang, S.;  et~al.
\newblock TotalSegmentator: robust segmentation of 104 anatomic structures in CT images.
\newblock {\em Radiology: Artificial Intelligence} {\bf 2023}, {\em 5},~e230024.

\bibitem[Heinrich et~al.(2013)Heinrich, Jenkinson, Brady, and Schnabel]{heinrich2013mrf}
Heinrich, M.P.; Jenkinson, M.; Brady, M.; Schnabel, J.A.
\newblock MRF-based deformable registration and ventilation estimation of lung CT.
\newblock {\em IEEE transactions on medical imaging} {\bf 2013}, {\em 32},~1239--1248.

\bibitem[Modat et~al.(2010)Modat, Ridgway, Taylor, Lehmann, Barnes, Hawkes, Fox, and Ourselin]{modat2010fast}
Modat, M.; Ridgway, G.R.; Taylor, Z.A.; Lehmann, M.; Barnes, J.; Hawkes, D.J.; Fox, N.C.; Ourselin, S.
\newblock Fast free-form deformation using graphics processing units.
\newblock {\em Computer methods and programs in biomedicine} {\bf 2010}, {\em 98},~278--284.

\bibitem[Mok and Chung(2020)]{mok2020large}
Mok, T.C.; Chung, A.C.
\newblock Large deformation diffeomorphic image registration with laplacian pyramid networks.
\newblock In Proceedings of the International Conference on Medical Image Computing and Computer-Assisted Intervention. Springer,  2020, pp. 211--221.

\bibitem[Tian et~al.(2024)Tian, Greer, Kwitt, Vialard, San Jos{\'e}~Est{\'e}par, Bouix, Rushmore, and Niethammer]{tian2024unigradicon}
Tian, L.; Greer, H.; Kwitt, R.; Vialard, F.X.; San Jos{\'e}~Est{\'e}par, R.; Bouix, S.; Rushmore, R.; Niethammer, M.
\newblock unigradicon: A foundation model for medical image registration.
\newblock In Proceedings of the International Conference on Medical Image Computing and Computer-Assisted Intervention. Springer,  2024, pp. 749--760.

\bibitem[Tian et~al.(2023)Tian, Greer, Vialard, Kwitt, Est{\'e}par, Rushmore, Makris, Bouix, and Niethammer]{tian2023gradicon}
Tian, L.; Greer, H.; Vialard, F.X.; Kwitt, R.; Est{\'e}par, R.S.J.; Rushmore, R.J.; Makris, N.; Bouix, S.; Niethammer, M.
\newblock Gradicon: Approximate diffeomorphisms via gradient inverse consistency.
\newblock In Proceedings of the Proceedings of the IEEE/CVF Conference on Computer Vision and Pattern Recognition,  2023, pp. 18084--18094.

\end{thebibliography}

\PublishersNote{}
\end{adjustwidth}
\end{document}